\documentclass[runningheads]{llncs}
\usepackage[T1]{fontenc}
\usepackage{graphicx}
\usepackage{hyperref}
\usepackage{multicol}

\begin{document}

\title{Generation of Diverse and Functional Robot Designs using Superquadrics Parametrisation and Quality-Diversity}

\author{Leni K. Le Goff\inst{1}\orcidID{0000-0003-1749-9154} \and
Sim\'on C. Smith\inst{1}\orcidID{0000-0001-5453-9659} \and
Emma Hart\inst{1}\orcidID{000-0002-5405-4413}}
\authorrunning{L. K. Le Goff et al.}
\titlerunning{Generation of Diverse and Func. Robot using SQ Parametrisation and QD}
%
\institute{Edinburgh Napier University, Scotland UK\\
\email{\{l.legoff2, s.smith2, e.hart\}@napier.ac.uk}}

\maketitle

\begin{abstract}
Generative design of robots requires navigating a vast search-space, encompassing physical configurations and behavioural
parameters.  Evolutionary Algorithms (EAs) have shown  promising results, but often converge prematurely to a small set of sub-optimal designs.
To counter premature convergence, we introduce a \textit{superquadrics}-based representation (SQs) for robot
bodies. SQs are interpretable, compact and computationally efficient
mathematical representations of 3D geometrical shapes that can be tuned to specific design-spaces.  To encourage morphological diversity, we combine this representation with a quality-diversity (QD)
algorithm (MAP-Elites). We compare SQs and Compositional Pattern Producing Networks representations as generators of morphologies, combining them with standard EAs and MAP-Elites. In two test environments, we find that using SQs to generate morphology in conjunction with the MAP-Elites algorithm reaches
the highest QD-score across both environments, maximising diversity of design and functionality of generated robots. The findings highlight the benefits of
using a compact and interpretable geometric representation for exploring a complex design-space and suggest that combining SQs with an explicit diversity mechanism increases the quality and number of 
designs generated.


\end{abstract}



\begin{figure}
  \includegraphics[width=\textwidth]{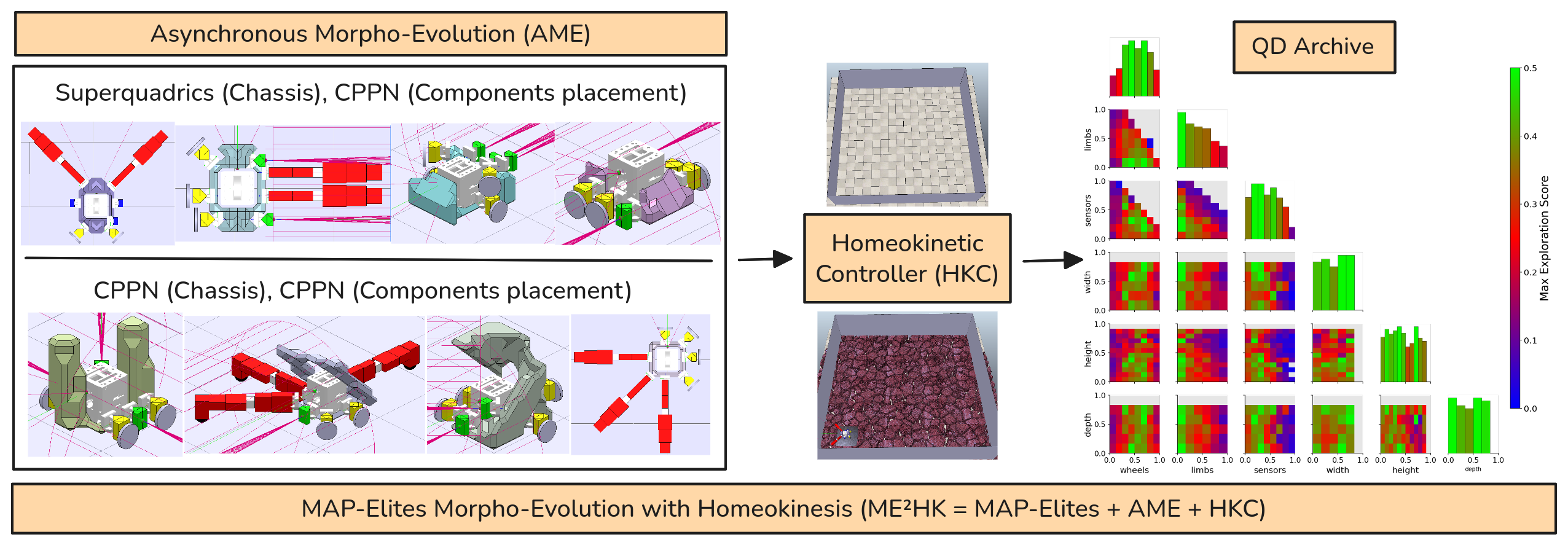}
  \caption{$ME^2HK^S$: 1) Asynchronous Morpho-Evolution, EA to optimise robot designs; 2) a superquadrics-based encoding for the robot chassis (compared to CPPN); 3) Homeokinetic controller, a self-organised behaviour generator; 4) MAP-Elites for diverse designs. 
  The output of $ME^2HK^S$ is an archive of functional and diverse robot.} 
  \label{fig:teaser}
  \vspace{-0.5cm}
\end{figure}
\section{Introduction}

Automatic generation of a robot is a difficult problem, comprising the simultaneous exploration and optimisation of the shape of a 3-dimensional object and its controller. Evolutionary algorithms (EAs) that jointly optimise design and controller have been shown to be a good fit to this problem due to their black-box nature. EAs can find optimal solutions even on a differentiable parametrisation of complex robot-design-space, with successful applications across a variety of robotic platforms, including modular/non-modular, soft robotics and 3D printed components~\cite{cheney2014unshackling,mouret2015illuminating,cheney2018scalable,kriegman2020scalable,li2023evaluation}.
Due to the complexity and high-dimensionality of search spaces, EAs can be computationally expensive, not allowing generation and testing of a large numbers of designs in a feasible time-frame~\cite{angus2023practical}. The low number of designs generated impacts the diversity of solutions explored, resulting in convergence to sub-optimal or simplistic designs~\cite{mertan2024investigating}. Achieving diversity in the design space is important for downstream tasks, where the most appropriate design is selected  and fine-tuned. For these reasons, all designs must be \textit{functional}, i.e. capable of solving, with a predefined performance level, a given task (different from the downstream one) in a arbitrary environment. Our goal is to design a method that maximises the diversity of functional robot designs to facilitate downstream optimisation.


For any EA, the representation (encoding) defines the size of the search-space and influences its ability to navigate that space. Algorithm designers try to simplify the search-space.
Several EAs for generative robot design use Compositional Pattern Producing Network (CPPN)~\cite{stanley2007compositional}. CPPNs are neural networks that produce diverse spatial patterns. These networks are well suited for designing morphologies which tend to have repetitive patterns, e.g. symmetrical bodies.
However, studies show issues with CPPNs such as bias toward certain type of designs and the difficulty of tuning its hyperparameters~\cite{thomson2025understanding}. 

To address the size of the search-space and bias in the representation, we propose a new encoding to generate a robot chassis based on \textit{superquadrics (SQ)}: arising from the field of Computer Vision, SQ is a compact and computationally efficient mathematical representation of 3D geometrical shapes that provides high representation strength, i.e., is able to accurately represent most common object types, using only 8 parameters~\cite{han2023sq}. 
SQs are less expressive than CPPNs due to their smaller parameter-space, but they are (a) more interpretable than a CPPN thanks to its parameters with ranges of values that correspond to specific shapes; (b) easier to optimise as they do not require a complex EA such as NEAT to evolve; (c) SQs are easy to tune for a specific design-space.


To increase diversity in the set of generated designs, we integrate the proposed superquadrics representation  with a recently proposed algorithm called Morpho-Evolution with Homeokinesis ($MEHK$) \cite{LeGoffICRA24}. This is a nested evolutionary algorithm in which an \textit{outer} loop evolves a design: we adapt this algorithm to utilise a  superquadrics representation to evolve the chassis design coupled with a CPPN representation to place actuators and/or sensors on the evolved chassis\footnote{placement of components via a CPPN is common in the literature e.g. \cite{li2023evaluation,LeGoffICRA24}}. For each design, an \textit{inner} loop learns a controller.  $MEHK$ replaces the learning algorithm that creates the controller with a  self-organised homeokinetic controller (HKC) \cite{homeok,hesse2009self}. The HKC can quickly  find a controller that enables any robot design to move in a single episode in a closed-loop fashion, therefore offering  a good proxy of the  potential functionality of the design. It has  been shown to be considerably faster the typical learning processes used in  morpho-evolution algorithms to create functional designs: for instance, in \cite{LeGoffICRA24} it was shown that  $MEHK$ can generate and evaluate 10000 designs in 40 hours using 64 CPUs, contrasting to e.g.  \cite{gupta2021embodied} that uses a deep reinforcement learning method requiring 1152 CPUs to generate only 4000 designs.

The $MEHK$ algorithm implicitly results in diversity by exploring a large space of robot designs. However, given that the family of evolutionary algorithms known as quality-diversity algorithms \textit{explicitly} generate diversity, we also propose to augment $MEHK$ with the MAP-Elites algorithm \cite{mouret2015illuminating} that imposes the EA population onto a multi-dimensional archive, with dimensions defined by morphological features.  By adding new solutions to the archive, the population grows over time, with solutions from the archive providing stepping stones to new areas of the search space \cite{nordmoen2021map,Nadizar25}. The contributions of the paper are twofold:
\begin{itemize}
    \item We propose the first use of the superquadrics representation within a nested morpho-evolution algorithm to improve chassis diversity, combining this with a CPPN to place components and a homeokinetic control mechanism to efficiently evaluate the functionality of robots;
    \item To increase the morphological diversity of the generated functional designs, we 
    adapt the previously proposed $MEHK$ algorithm to function as  quality-diversity algorithm (MAP-Elites), using a 6-dimensional archive describing features of the design.
    
    
\end{itemize}

Experiments are conducted 
using two representations (SQ-CPPN, Dual CPPN, see Section~\ref{sec:enc}) in conjunction with two algorithms: $MEHK$ using an elitist selection mechanism following~\cite{LeGoffICRA24}  and a version of $MEHK$ that uses MAP-Elites to project individuals into a 6-dimensional archive describing  morphological characteristics, 
and in which the selection mechanism randomly chooses solutions from the archive, following \cite{mouret2015illuminating}.  We find that the SQ-CPPN representation used in conjunction with MAP-Elites increases both the quality and number of 
designs generated.


In the remainder of the paper: $MEHK^C$ and $MEHK^S$ refer to using $MEHK$ from \cite{LeGoffICRA24}, where $(C,S)$ respectively refers to the Dual CPPN and SQ-CPPN representation, while $ME^2HK^C$ and $ME^2HK^S$ refer to the use of $MEHK$ in conjunction with a MAP-Elites algorithm.



\section{Related Work}



The choice of representation, as a direct or indirect encoding of a robot design, defines the size of the search space and the ease by which it can be navigated. Veenstra \textit{et al.} \cite{VeenstraEncodings} show that the type of encoding is at least as important for creating robots as the optimisation strategy used, corroborating earlier findings in voxel-based soft robots \cite{medvet2021biodiversity}. 

In modular robotics, it is common to use an indirect encoding of a design, for example L-Systems \cite{lindenmayer1992grammars} or graph-rewrite approaches \cite{miras2018effects}. A common indirect approach is to use a Compositional Pattern Producing Network (CPPN) \cite{stanley2007compositional} that learns a mapping between spatial coordinates and the module or material that should be placed at each coordinate. CPPNs generate structured patterns that are regular, symmetric, and repetitive.
These networks have been used in evolutionary modular robotics \cite{kriegman2020scalable,auerbach2011,mouret2015illuminating}. In \cite{li2023evaluation,legoff2024efficient}, the authors evolve free-form chassis shapes for a robot using CPPNs. However, CPPNs often lead to fast convergence to a specific type of robot. For example, Miras \textit{et al}. \cite{miras2021constrained} observe that when evolving modular robots, the optimisation process frequently produces cross-shape robots (dubbed `spiders') and dysfunctional robots that are incapable of moving. The authors of \cite{cheney2018scalable,mertan2024investigating,mertan2025evolutionary} show the tendency of  evolutionary joint optimisation processes to converge to sub-optimal robot designs, noting that typical design and controller  joint optimisation algorithms consistently fails to select morphological potential.  Thomson \textit{et al.} \cite{thomson2025understanding} study how algorithms navigate the search-space during joint optimisation using Local Optima Networks. The authors find that compared to an L-System, the use of CPPNs leads to significant bias in the designs with low-quality local optima.

\textit{Superquadrics} is a compact mathematical way of representing a 3D shape commonly used in robotics related to grasping and pose estimation \cite{vezzani2017grasping,wu2025autonomous}. Superquadrics belong to a family of parametric representations, and have been used in computer vision and engineering applications (e.g. design of mechanical components \cite{Dupac2012}).
To the best of our knowledge, the only example of superquadrics with EA is in \cite{husbands1996two} from 1996, where authors search the space of shapes for `interesting' designs of 3D objects in an interactive EA. In \cite{scarton23}, the authors evaluate multiple representations for designing architectural shapes. Their work includes evaluation of a parametric representation. However it is restricted to low-dimensionality encoding. Thus, not appropriate in a robotic context.

To increase \textit{diversity} in optimisation tasks, several authors use Quality-Diversity (QD) algorithms like MAP-Elites \cite{mouret2015illuminating}.
QD facilitate the exploration of the search-space and can generate a set of diverse designs for multiple downstream tasks.
QD search includes an organised archive whose axes are defined by multiple descriptors. These axes describe user-defined characteristics. The early applications of MAP-Elites opt to maximise \textit{behavioural} diversity.
However, descriptors can reflect  characteristics of the \textit{morphological} space. In~\cite{mouret2015illuminating}, a 2D descriptor space is defined by the percentage of voxels filled and the percentage of voxels that are made of `bone' material.  In \cite{xie2025morphology},  robot designs are mapped to features based on a number of morphological complexity metrics, including voxel heterogeneity, structural connectivity, symmetry, and actuator distribution. In \cite{xie2024map}, a 2D descriptor is used with MAP-Elites to quantify diversity arising from the configuration of `voids' in the morphological space, via Shannon entropy and void-centroid distance.



This paper builds on previous work in two directions. It  addresses issues  associated with the CPPN representation via  introducing \textit{superquadrics} to represent the design, and adapts the existing $MEHK$ algorithm to  explicitly maintain morphological  diversity using MAP-Elites.

\section{Methodology}

\subsection{Morpho-Evolution with Homeokinesis}
We repurpose the $MEHK$ method proposed in  \cite{LeGoffICRA24} to evolve a diverse set of robots, adapting it to the format of MAP-Elites in which a population is superimposed on an archive discretised into cells. $MEHK$ has two nested processes: (1) an outer process, called Asynchronous Morpho-Evolution (AME) that generates designs and  (2) for each design, an inner process, called Homeokinetic Control (HKC), that explores the sensorimotor space of the robot resulting in seemingly intelligent behaviour with minimal adaptation time (See appendix A for a diagram and pseudo-code of AME and appendix B for more information on HKC).  Hyperparameter values  for $MEHK$ are in appendix C.

\paragraph{Asynchronous Morpho-Evolution (AME).} 

AME is an asynchronous EA~\cite{le2024improving}: i.e.  a form of steady-state EA in which new offspring are generated as soon as one robot finishes its evaluation step. Asynchronous EAs can be efficiently parallelised as the production of new  offspring does not have to wait for an entire population to be evaluated. The algorithm manages two sets of individuals: a population and an evaluation queue. The population corresponds to the set of parents available to take part in reproduction. The evaluation queue corresponds to the set of robots waiting to be evaluated. The population is updated following three steps: (1) \textit{Replacement}, where newly evaluated individuals are added to the population and the worst individuals are then removed to keep the population constant size; (2) \textit{Selection}, where parents are selected using a tournament of size 4\footnote{This value is used in previous works using AME~\cite{le2024improving,LeGoffICRA24} }; (3) \textit{Reproduction}, where the selected parents are mutated to produce offspring which are added to the evaluation queue. At each update of the population, the number of offspring produced is equal to the number of individuals added to the population to keep the size of the evaluation queue constant.  We set the population and evaluation queue size to 100 as in  the original $MEHK$ paper~\cite{LeGoffICRA24}. The algorithm is initialised with 100 random robot designs. 

The fitness function used for AME measures the extent to which a robot is able to explore an environment, quantified by a \textit{exploration score}. The arenas shown in Figure~\ref{fig:environments} are divided into a grid of 16 by 16 tiles. The exploration score is computed by counting the number of tiles the robots visited at least once and normalised over the total number of tiles.

\paragraph{Asynchronous MAP-Elites Morpho-Evolution ($AME^2$).}

To augment AME with MAP-Elites, the unstructured population used in AME is replaced by a grid-based archive. The grid is multi-dimensional with each dimension corresponding to a discretised range of features of the robot designs (see Section~\ref{sec:rds}). 
Compared to AME, the \textit{replacement} and \textit{selection} steps are altered. A newly evaluated individual is added to the archive if the cell corresponding to its features is empty or if it is occupied and its fitness value is higher than the current stored design. To generate new offspring and fill the evaluation queue, individuals are randomly selected from the archive and mutated. The evaluation queue is kept at a constant size of 100, so, new offspring are generated until the queue is filled.

\paragraph{Self-organisation of behaviour.}

Homeokinesis is a concept arising in the field of physics from studying self-organisation and complex systems~\cite{soodak1978homeokinetics}.  Homeokinetic control algorithms has been applied to robotics to quickly adapt its internal parameters resulting in seemingly intelligent behaviours without specific goal~\cite{der2012playful,smith2011homeokinetic}. The principle of the homeokinetic controller (HKC) is to balance the \textit{predictability} and \textit{sensitivity} of the dynamics of the interaction between the robot and the environment. The behaviour of a robot is predictable if the resulting state of executing an action can be predicted by an internal forward model. HKC measures predictability by training the \textit{forward model} and computing the error between the predicted state and the state obtained after executing the action. Sensitivity is the change in the sensory signal after performing an action. Homeokinesis calculates the sensitivity of a dynamical system as the Jacobian of the state over the actions. An adaptation loop in HKC maximises sensitivity and predictability of the dynamics at every time step. The result is a coherent behaviour that emerges from the correlation between the degrees of freedom while maximising predictability and sensitivity. A robot that only maximises sensitivity will behave chaotically, while one that only maximises predictability will remain still. An exploratory behaviour arises in the balance between sensitivity and predictability at the edge of chaos. For example, a wheeled robot will excite its 2 wheels and make the robot move in lines, circles, or rotate on the spot. Furthermore, the robot will react to its surrounding, e.g. escaping from a non-sensitive state, like being stuck in corner. 
Note that homoekinesis is a quick adaptive algorithms, that do not require long training phases to generate the behaviours. 
The current joints angles, wheels positions and proximity sensors values are the input of the controller, and its outputs are used to command the joints angles and wheels angular velocities. Homeokinesis is agnostic to the configuration of the robot, is able to generate the coherent behaviour in any configuration while only requiring the input and output dimensions. We include the linear model implementation of Homeokinesis in the Supplementary Material.

\subsection{Robotic design space}\label{sec:rds}

\begin{figure}[ht]
\vspace{-0.8cm}
  \centering
  \includegraphics[width=0.7\linewidth]{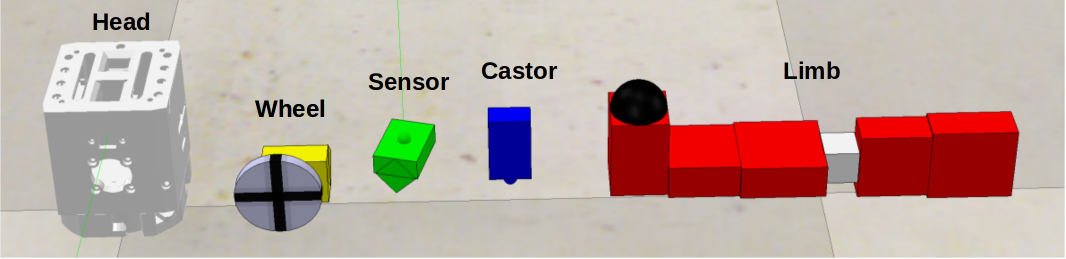} 
  \caption{Five components for the robot's design.}
  \label{fig:robot}
  \vspace{-1cm}
\end{figure}

\subsubsection{Design Space}
The robot designs are based on the ARE robotic platform \cite{buchanan2020bootstrapping,angus2023practical}. This platform utilises 5 hand-designed components (see Figure~\ref{fig:robot}) that can be assembled on a free-formed chassis (which is subject to evolution). The "head" is the central computing unit of the robot and is always placed in the bottom centre of the chassis.  Wheels, sensors, limbs and castors are assembled on the surface of the chassis. The component positions are decided by the genotype --- up to 8 components can be placed. The sensors are  proximity sensors that output a value between 0 and 1, while limbs are composed of two active hinges: the one attached to the chassis rotates horizontally and the second one rotates vertically. Robots designs are enclosed in an 11 entries squared 3D matrix where each voxel can contain a piece of chassis, limb, wheel, sensor, castor or nothing. 

For the $MEHK$ variants augmented with MAP-Elites, the grid archive has 6 dimensions: the number of wheels, joints, sensors and the height, depth and width of the chassis. The range and number of bins for each dimension are the following: \textit{wheels}, $\{0,8\}$, 8 - \textit{joints}, $\{0,12\}$, 6 -  \textit{sensors}, $\{0,8\}$, 8 - \textit{height}, $[0,1]$, 12 - \textit{depth}, $[0.5,1]$, 6 - \textit{Width}, $[0.5,1]$, 6.
         

\subsubsection{Encoding}\label{sec:enc}
In the experiments, we are comparing two encodings: SQ-CPPN and Dual CPPN. They are composed of two parts:  the shape of the chassis and the type and placement of the components. For the chassis, depending on the encoding used, a superquadrics representation (SQ) or a CPPN is used. For the components, a CPPN is always used. First, the chassis is generated by inputting the spatial coordinates of each voxel into the SQ equation or a CPPN. Then, another CPPN is queried for each coordinate of the surface of the chassis to determine if a component should be placed and if so, of which type.

\paragraph{Superquadrics (SQ)}
As previously noted, superquadrics are a family of geometric shapes that can be fully described using 8 parameters (see equation~\ref{eq:quad}).  The equation~\ref{eq:quad} is the "in-out" formula of a superquadric which defines a \textit{surface} with equation $Q_n(x,y,z) = 1$ or a \textit{volume} with inequality $Q_n(x,y,z) \leq 1$.  If the value of $Q_n(x,y,z)$ is less or equal to one, then a piece of chassis is added at the voxel of this coordinate.

\begin{figure}[h!]
\vspace{-0.5cm}
\begin{minipage}{0.5\textwidth}
    \centering
    \includegraphics[width=0.7\linewidth]{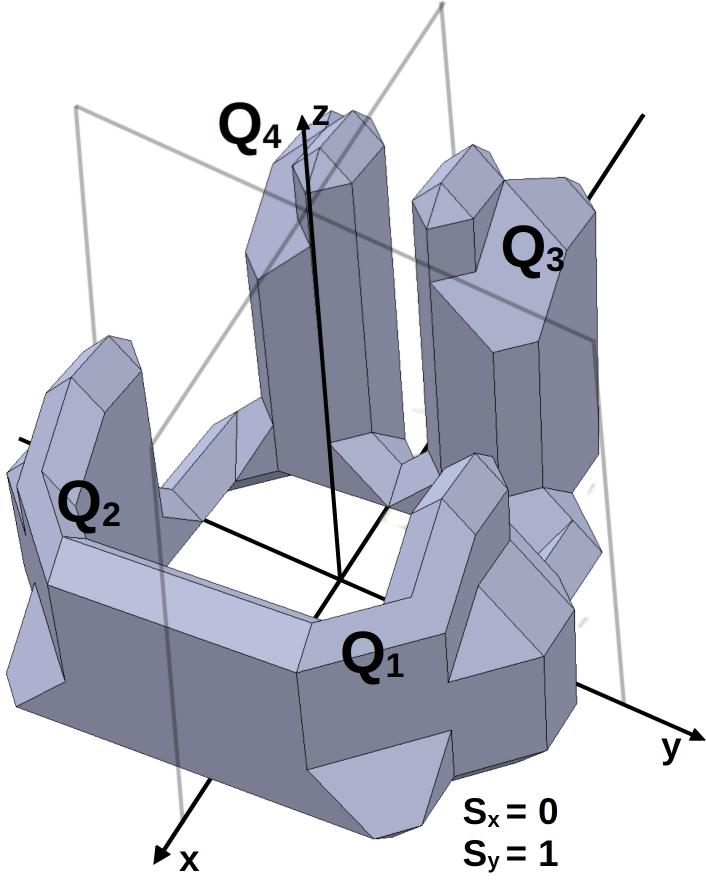}
    \caption{Example of a chassis generated via superquadrics, with the 4 quarters shown. This chassis was generated with $S_x = 1$ and $S_y = 0$, so, has symmetry on the x-axis.}
    \label{fig:chassis_sq}
\end{minipage}
\begin{minipage}{0.5\textwidth}
  \centering
  \includegraphics[width=0.49\linewidth]{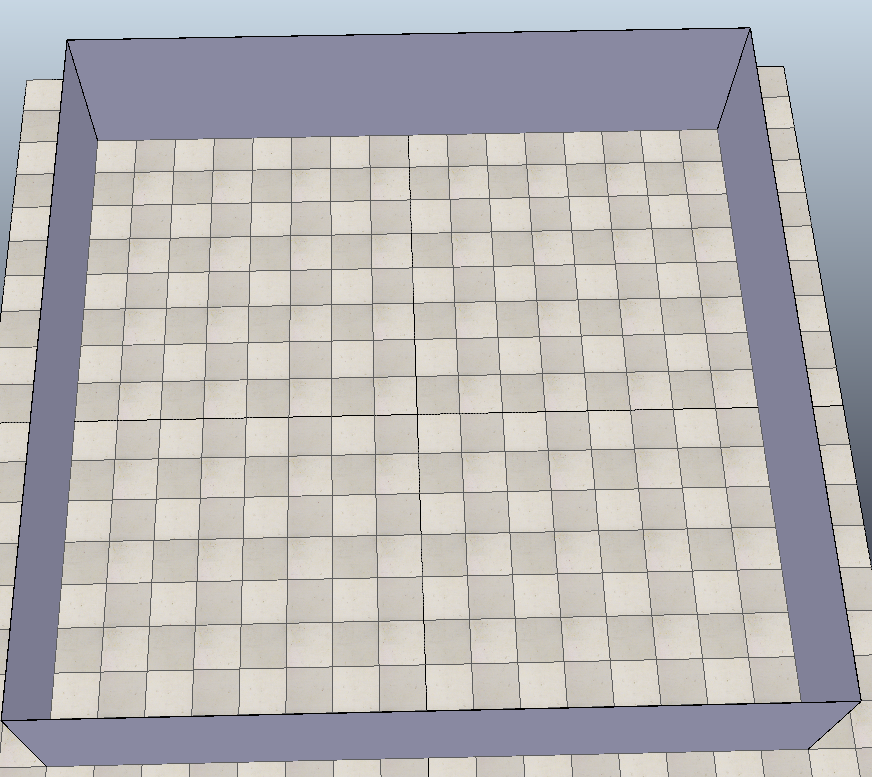}
  \includegraphics[width=0.49\linewidth]{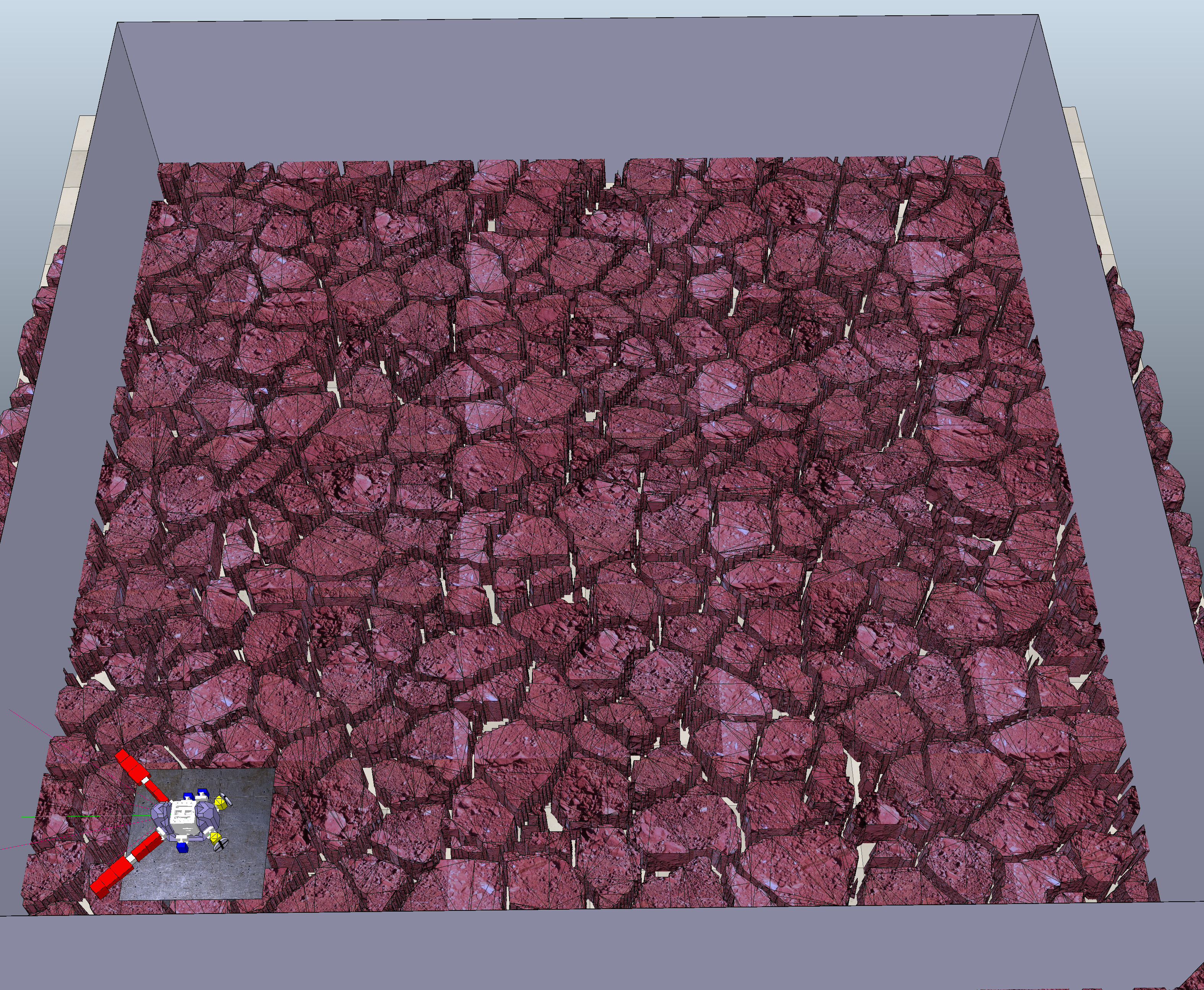}
  \caption{The two environments used in the experiments. From left to right image: flat and cracks. All the environments are 4 by 4 metres. A robot is included in the picture of the cracks environment, giving an idea of the relative size of the robots in respect to the environment.}
  \label{fig:environments}
\end{minipage}
    \vspace{-0.5cm}

\end{figure}

To increase the variety of shapes that can be generated, 4 superquadrics are used. Each superquadric corresponds to a quarter of the possible space that the chassis can occupy: $Q_1 : (x\ and\ y) \geq 0$, $Q_2: x \geq 0\ and\ y < 0$, $Q_3 : x < 0\ and\ y \geq 0$, and $Q_4: (x\ and\ y) < 0$ (see Figure~\ref{fig:chassis_sq}). The z coordinate is always positive which guarantees a flat base to the shape.
\begin{equation}
    Q_n(x,y,z) = \left(R\left(\frac{|x|}{A}\right)^{\frac{2}{u}} + S\left(\frac{|y|}{B}\right)^\frac{2}{u}\right)^\frac{u}{v} + T\left(\frac{|z|}{C}\right)^{\frac{2}{v}}
    \label{eq:quad}
\end{equation}
{\small
\textbf{ Equation 1:   $x,y,z$ are 3d coordinates, $A,B,C,R,S$ and $T$ are free parameters changing the proportion and size of the shape and their sign change the type of shape. $u$ and $v$ are free parameters changing the type of shape. For some set of parameters, the equation is not mathematically valid if there are negative coordinates, so the absolute function is applied to the inputs (x,y,z).\newline
}}

Equation~\ref{eq:quad} gives a meaningful parametrization of the superquadrics. The range of values of these parameters can be mapped to the proportion, size and type of shape. Two binary parameters, $S_x$ and $S_y$, determine if the shape is symmetric on the \textit{x} or \textit{y} axis. If $S_x = 1$ then $Q_1 = Q_2$ and $Q_3 = Q_4$; if $S_y = 1$  then $Q_1 = Q_3$ and $Q_2 = Q_4$. 
The 32 parameters of the superquadrics are mutated using Gaussian mutation; $S_x$ and $S_y$ are randomly chosen at the initialisation and not mutated during evolution. So, depending on the value of $S_x$ and $S_y$ the genotype has a fixed size of 10, 26 or 34 throughout the evolution.  To ensure that AME starts with mathematically valid superquadrics, the 8 parameters of $Q_n$ are picked randomly from 9 sets of values where each set corresponds to a type of shape (see appendix C).

\paragraph{Compositional Producing Pattern Network (CPPN)}


The CPPN takes three inputs describing the spatial coordinates of the voxel and outputs one value between -1 and 1. When a CPPN is used to encode the chassis, a positive value means a piece chassis is added, otherwise a negative value means nothing is added. Using a CPPN to generate the chassis can lead to the creation of pieces that are not attached to the body, i.e are floating: such pieces are discarded. Note that floating pieces do not happen with SQ.

To decide the placement and type of components, another CPPN is used. Its output interval $[-1,1]$ is split in 5 equal intervals corresponding to each component type: $[-1,-0.6[$ for no component, $[-0.6,0.2[$ for wheel,$[-0.2,0.2[$ for sensor, $[0.2,0.6[$ for joint, and $[0.6,1]$ for castor. Finally, a \textit{manufacturability} check ensures the initial design can be simulated and built. The check ensures that there are no collisions between components or with the chassis and reduces the number of components to a maximum of 8\footnote{The maximum of 8 components is a constraint introduced when building physical robots due to power requirements \cite{angus2023practical}}. If  more than 8 components remain  after the collision check, the 8 closest to the ground are kept.

Both CPPNs are mutated using the Direct-encoding Topology Evolution~\cite{le2020pros} (DET) algorithm. DET is a neuro-evolution algorithm that optimises the topology, weights and biases of the CPPN using 7 mutation operators: adding or removing a node, adding, removing or changing a connection, changing the weights of the connections and the biases of the nodes using Gaussian mutation (see Appendix C for the mutation parameters). In this paper, DET is combined with AME. For the $MEHK$ variants using CPPN to encode the chassis, two CPPNs are evolved: one to generate the chassis and another to place and select the components. In the rest of the paper, we refer to this encoding as \textit{Dual CPPN}. In the variants using SQ, only one CPPN is used for the components and the SQ to generate the chassis. We will refer to this encoding as \textit{SQ-CPPN}

\subsection{Experimental Protocol}


We compare four different variants of $MEHK$: $MEHK^C$, $ME^2HK^C$, $MEHK^S$, and $ME^2HK^S$. The superscripts (C,S) indicates respectively the use of Dual CPPN or SQ-CPPN. The square symbol indicates use of MAP-Elites. Each of these variants is run in two different environments (Figure~\ref{fig:environments}), each of which is designed to target particular features of the design space. The \textit{flat} arena would be expected to favour robot design with wheels,
while navigating the \textit{cracks} terrain  might be expected to favour limbs that could prevent the robot from getting stuck.


For each environment, experiments are replicated 20 times. The budget of a run is the evaluation of 10000 designs. Each new robot designed has 20 minutes to explore the arena using HKC. These parameters are taken from  \cite{legoff2024efficient}.
The variants of $MEHK$ are compared using the following metrics: (1) the \textit{archive coverage} measures the number of cells filled in the 6-dimensional grid archive; (2) the \textit{Quality-Diversity (QD) score} is the sum of the fitness values of all the robots stored in the archive; and (3) the maximum \textit{exploration or fitness score}. To compute the archive coverage and QD score for the $MEHK$ variants without MAP-Elites,  at the end of the evolutionary process the 10000 robots generated  are assigned to the same 6-dimensional grid as used in MAP-Elites.

The source code used to run the experiments and supplementary materials with the appendices are available on \href{https://doi.org/10.5281/zenodo.19631049}{Zenodo}.

\section{Results}

\begin{figure*}[t]
    \centering
    \includegraphics[width=\linewidth]{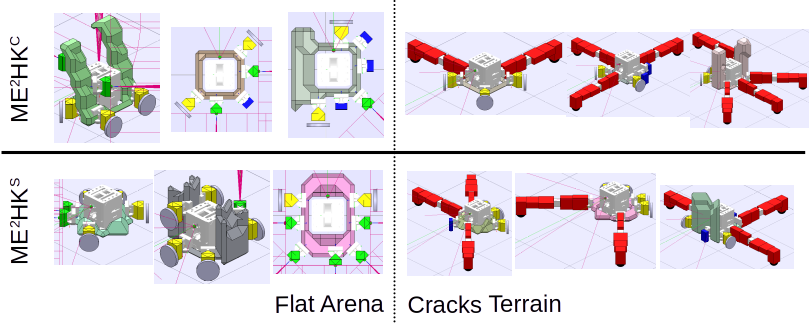}
    \caption{Pictures of robots picked from the top 10 robots in term of exploration score for $ME^2HK^C$ and $ME^2HK^S$. See appendix E for more pictures. }
    \label{fig:robex}
    \vspace{-0.5cm}
\end{figure*}

\begin{figure}[h!]
    \centering
    \includegraphics[width=0.4\linewidth]{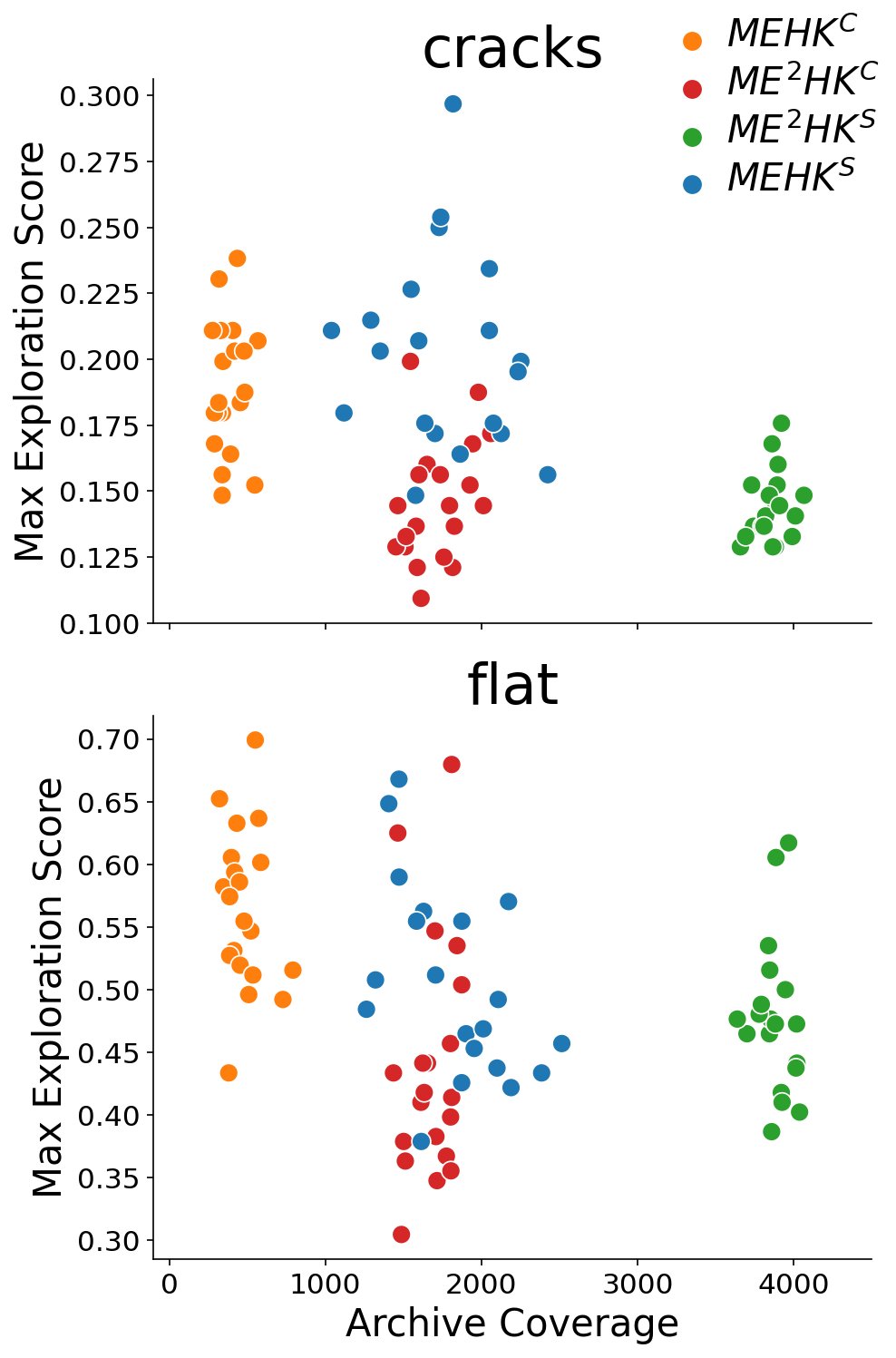}
    \includegraphics[width=0.4\linewidth]{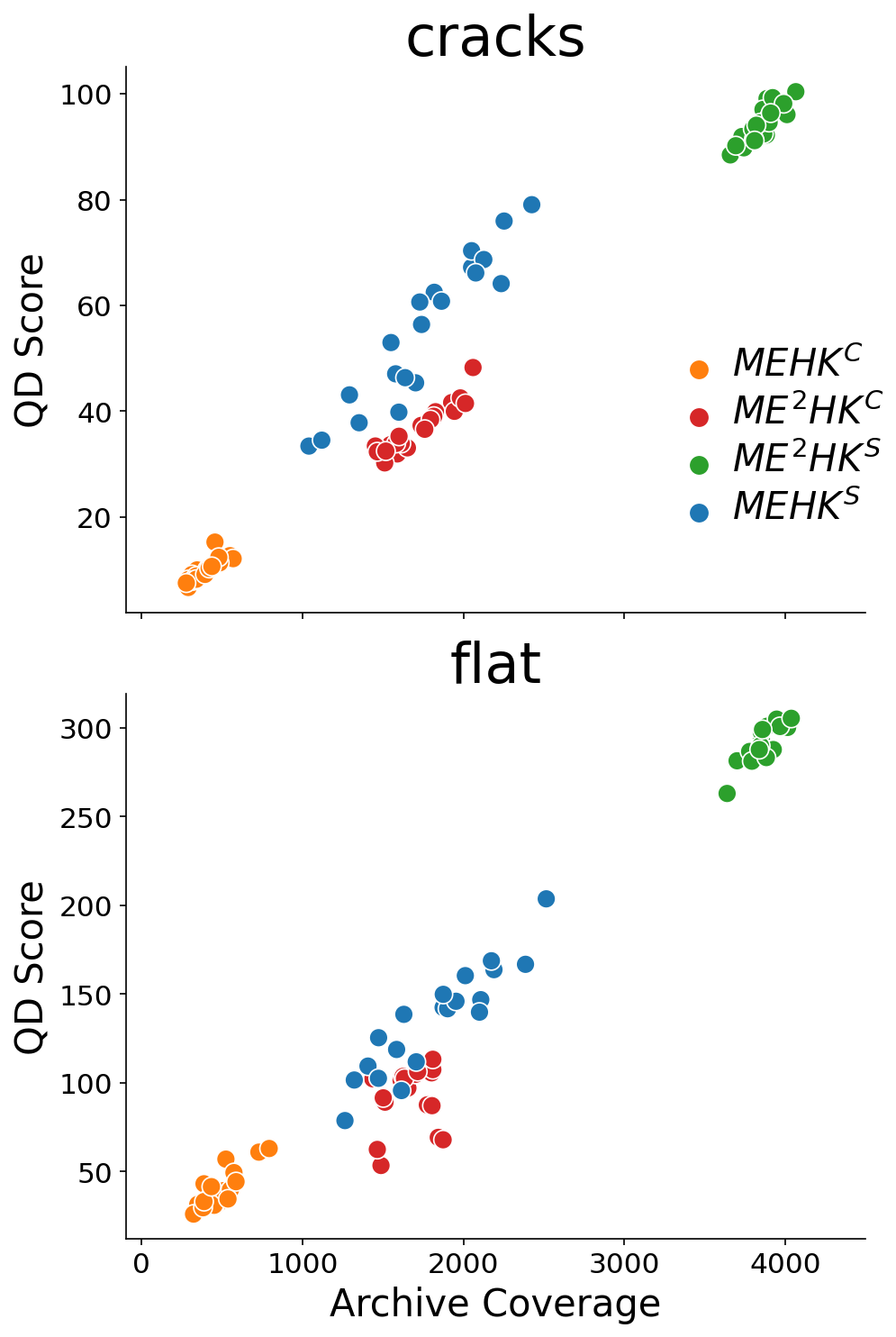}
    \caption{Distribution of the replicates over the max exploration score and archive coverage.}
    \label{fig:covfit}
    \vspace{-0.7cm}
\end{figure}

Overall, the best variant of $MEHK$ is $ME^2HK^S$ reaching the highest archive coverage and QD score (statistically significant) for both environments (see table~\ref{tab:tabres} and Figure~\ref{fig:covfit}). Using the SQ-CPPN encoding, $MEHK$ generates \textit{four times} more robot designs stored in the archive than when using Dual CPPNs encoding.  This number is again doubled when adding MAP-Elites. When adding MAP-Elites on top of Dual CPPNs encoding, the archive coverage is quadrupled. As expected, regardless of the encoding, MAP-Elites increases greatly the diversity of solutions explored. 
   
\begin{table}[h]
    \centering
    \small
    \begin{tabular}{c|c|c|c|c|}
      & \multicolumn{4}{c|}{$MEHK$}      \\ \hline
  & \multicolumn{2}{c}{Dual CPPNs}        & \multicolumn{2}{|c|}{SQ-CPPN}      \\ \hline
      & Flat Arena       &Cracks  &Flat Arena        &Cracks\\ \hline
AC      & $483.5 | 120.8^{**}$      & $384.3 | 86.8^{**}$      &$1827.3 | 356^*$    &$1761.2 | 381.5^*$\\ 
QD               &$39.4 | 10.7^{**}$        & $9.74 |  2.15^{**}$   & $135.7 | 30.2^{**}$     & $55.6 | 14^{**}$\\
Fit$^M$     & $\mathbf{0.56 | 0.06^*}$    & $0.19 | 0.03^*$   & $\mathbf{0.5 | 0.08^*}$            & $\mathbf{0.2 | 0.04^*}$\\ 
\hline
      & \multicolumn{4}{c|}{$ME^2HK$}      \\
\hline
        &\multicolumn{2}{c}{Dual CPPNs}        & \multicolumn{2}{|c|}{SQ-CPPN} \\     
\hline
       &Flat Arena        &Cracks  & Flat Arena      &Cracks\\
  C   & $1678.2 | 139.3^*$         & $1718.6 | 193.1^*$ & $\mathbf{3880.6 | 105.7^{**}}$      &$\mathbf{3854 | 103.1^{**}}$\\
QD        &$92.6 | 17.1^{**}$     & $37.8 |   4.6^{**}$    &$\mathbf{292 | 10.4^{**}}$       &$\mathbf{94.3 | 3.4^{**}}$\\
Fit$^M$     &$0.44 | 0.1^*$      &$0.15 | 0.023^*$   & $0.48 | 0.08^*$       &$0.14 | 0.013^*$\\
\hline
    \end{tabular}
    \caption[This is the caption]{Summary metrics: AC=archive coverage; QD=QD Score; Fit$^M$ = max fitness score, showing
    [\textit{average} | \textit{standard deviation}] for each experiment.
    The values in bold corresponds to the highest value overall. The stars indicates if the differences are statistically significant\footnotemark.
    }
    \label{tab:tabres}
    \vspace{-0.5cm}

\end{table}

\footnotetext{** indicates significant difference and large effect size with all the other variants and * indicates significant difference and large effect size with all but one other variant. Statistical difference is computed with Mann-Whitney U test ($p <  0.001$) and the effect size using Cliffs delta considered large difference ($\delta \ge 0.5$).  
}

The maximum fitness values or exploration score for all variants are similar (no statistically significant difference) with around 50\% of the flat arena and between 14-19\% of the cracks arena visited. Therefore, all the variants are able to produce robots with similar level of mobility. This result is further supported by the plots in Figure~\ref{fig:covfit} showing a similar spread of exploration scores across the replicates for all the variants and both environments. Moreover, on the right hand side plots of Figure~\ref{fig:covfit}, the QD score increases proportionally with the archive coverage, showing that the quality of the robots added to the archive is maintained. In other words, $ME^2HK^S$ generates additional robots with similar mobility as the ones generated by the other variants.

\begin{figure*}[h!]
    \vspace{-0.2cm}
    \centering
    \includegraphics[width=0.45\linewidth]{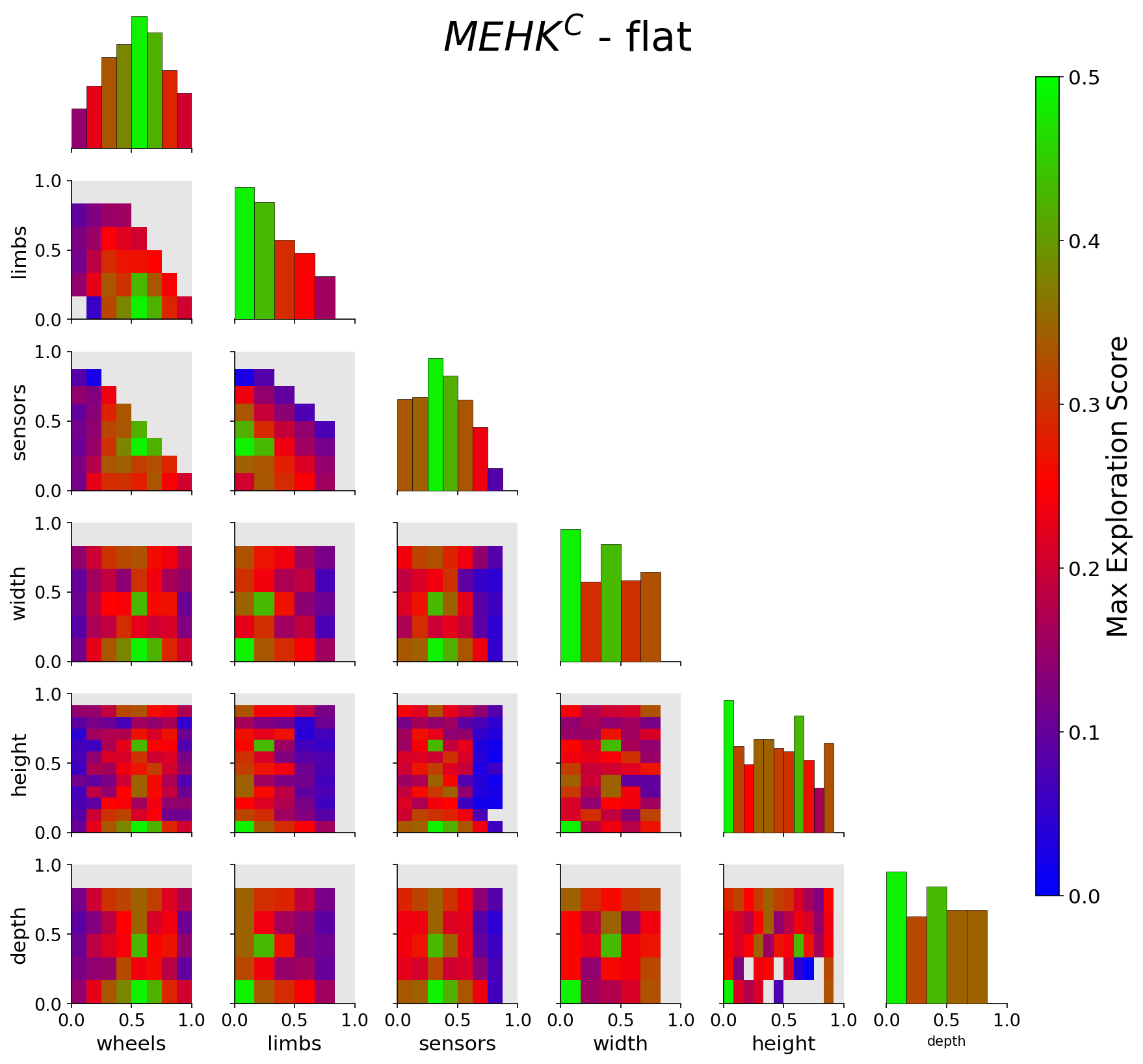}
    \includegraphics[width=0.45\linewidth]{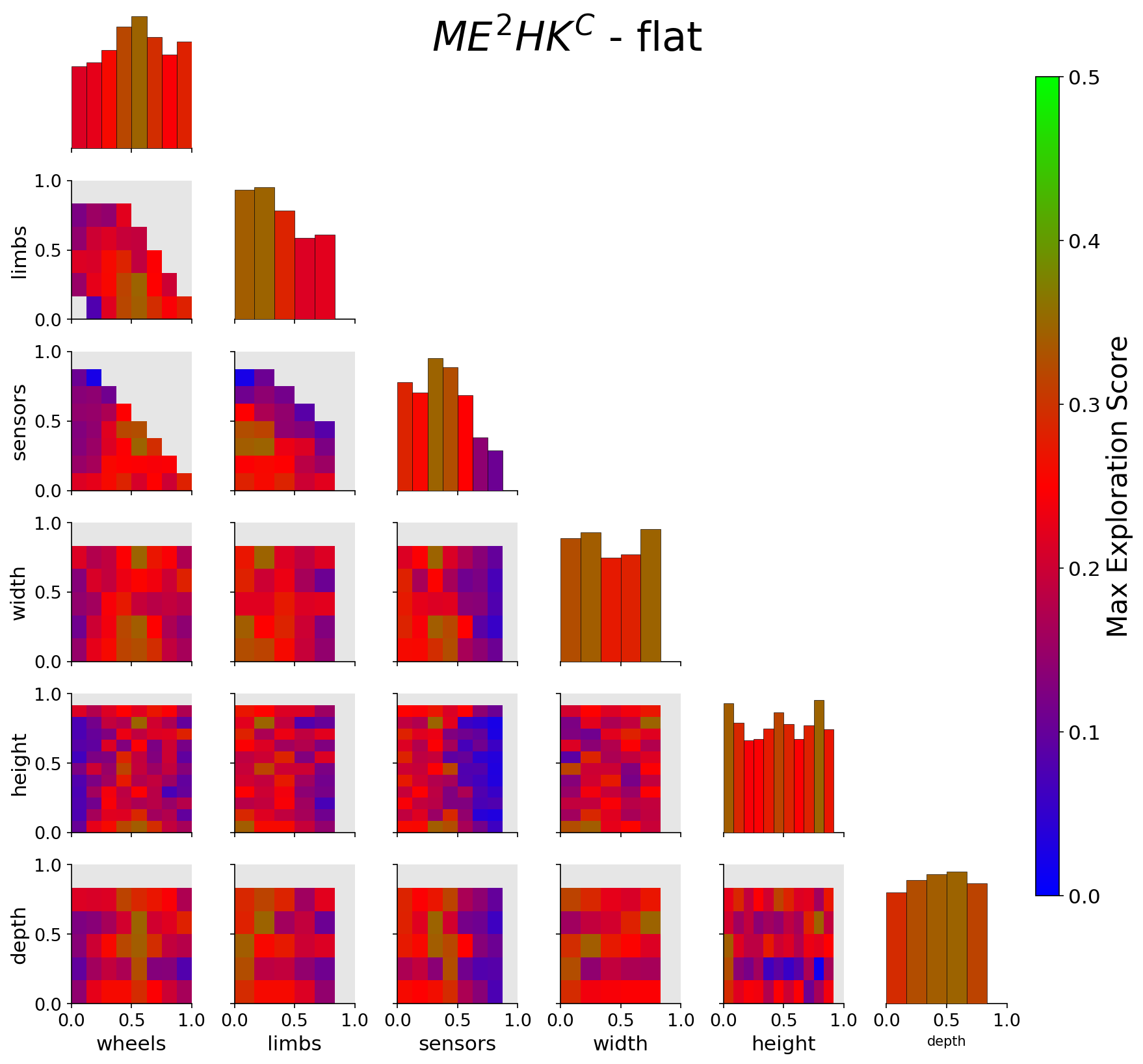} \\
    \includegraphics[width=0.45\linewidth]{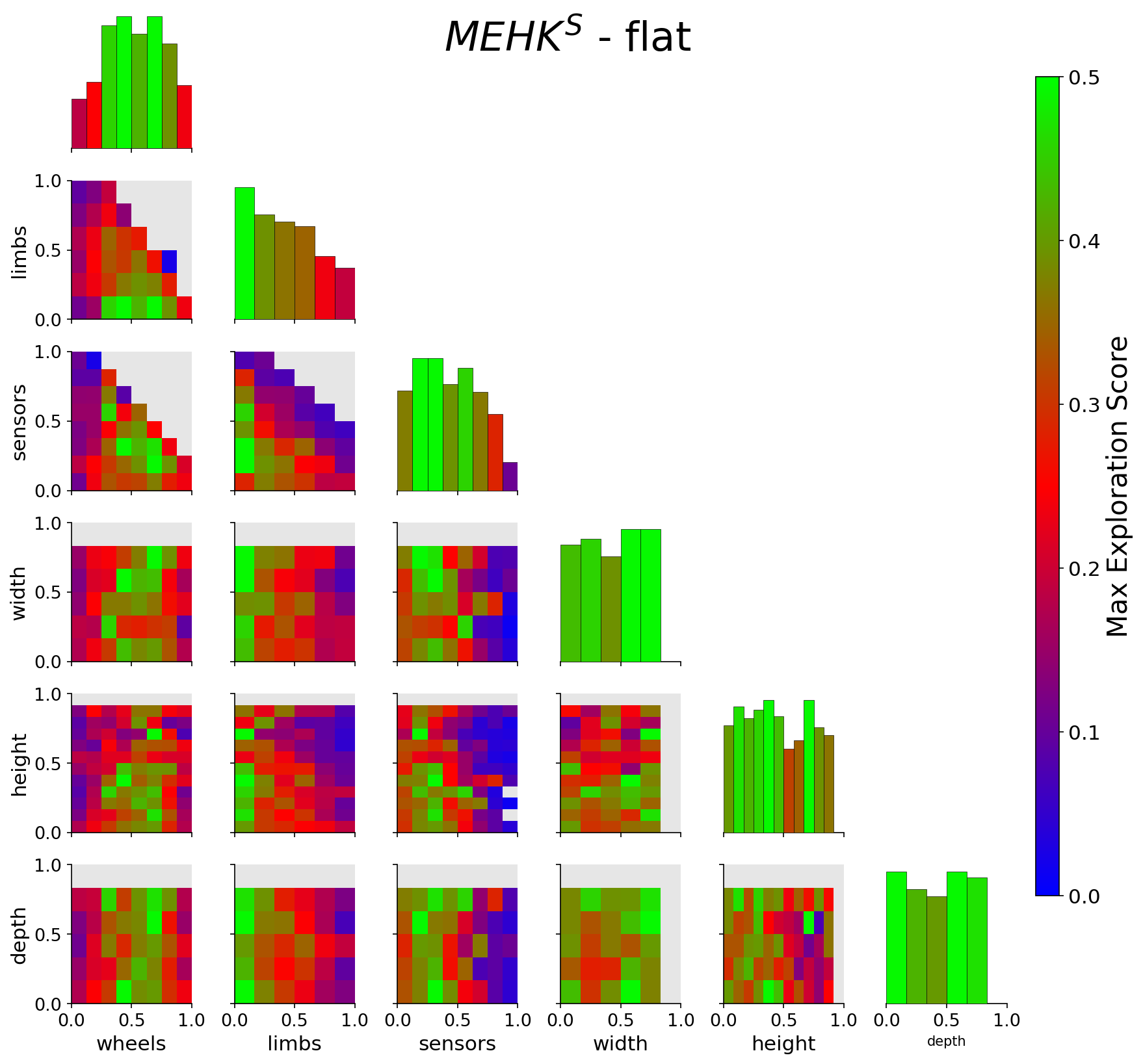}
    \includegraphics[width=0.45\linewidth]{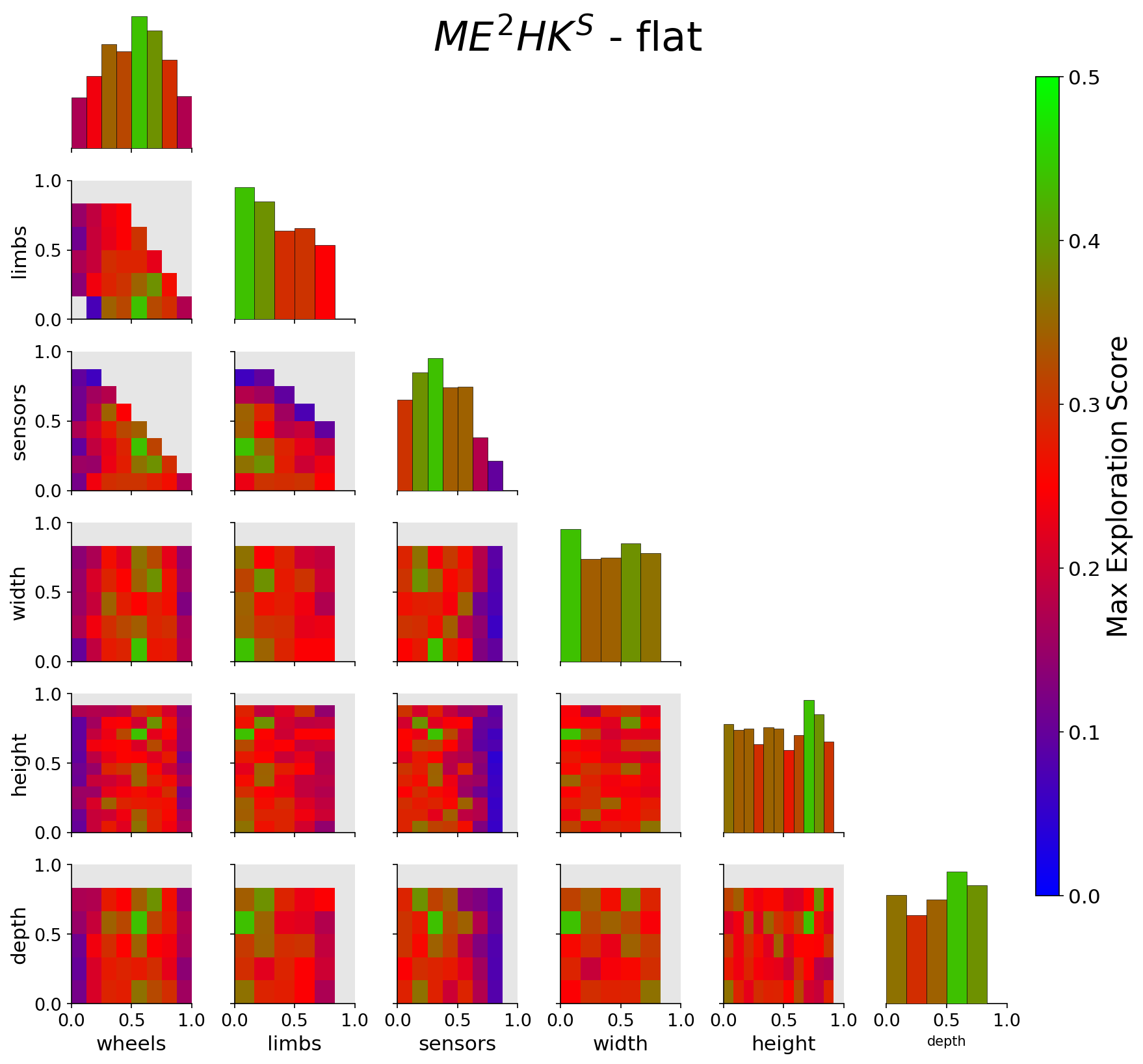}
    \caption{Grid archive pair plot for each variant on the flat arena. The heat-maps show the projection of the archive on 2D and the histograms in the diagonal show the projection on one dimension, length of the bars is proportional to exploration score. Colours show exploration score: blue as 0\%, red 25\% and green 50\% of the arena explored. }
    \label{fig:flat_archive}
     \vspace{-0.5cm}
\end{figure*}

Figure~\ref{fig:flat_archive} visualises the archives populated by each variant in the flat arena. The archives of all the replicates are aggregated in this figure, while the exploration score is  the maximum exploration score averaged over the replicates.  The use of MAP-Elites helps to increase both the diversity and quality of solutions. The archive of $ME^2HK$ has fewer unfilled cells than $MEHK$ regardless of the encoding. Robots with an exploration score around 0.25 (red colour) are more spread in the archives of $ME^2HK$ regardless of the encoding. However, there are fewer robots with exploration scores around 0.5 (green colour). This result suggests that MAP-Elites with $MEHK$ spreads the optimisation effort over a larger number of robot designs while $MEHK$ focuses on fewer designs. As a consequence, $ME^2HK$ produces fewer high performing robots than $MEHK$ but increases the performance of a larger set of unique robot designs. In particular, designs with the highest number of limbs (6) and sensors (8) have a higher exploration score with MAP-Elites, suggesting that they were further optimised resulting in a better configuration of components. 


Figure~\ref{fig:flat_archive} shows that $MEHK$ using SQ-CPPN (with or without MAP-Elites) increases the overall quality of the designs in comparison with using Dual CPPN. There are fewer designs with an exploration score between 0.4 and 0.5 (green colour) with Dual CPPN than with SQ-CPPN.  For $MEHK^C$, the designs with a high fitness, between 0.4 and 0.5, are not spread over the archives. The best robot designs have a low number of limbs and a high number of wheels and sensors. 
With respect to the chassis design, the width, height and depth have either small values or values around 0.5 (0.7 for the height).
With SQ-CPPN, these high performing designs are more predominant in the archive, particularly for the chassis dimensions (as expected considering that SQ encodes the chassis). The superquadrics-based encoding alone improves the diversity and quality of solutions.

\begin{figure}[h!]
    \vspace{-0.3cm}
    \centering
    \includegraphics[width=0.48\linewidth]{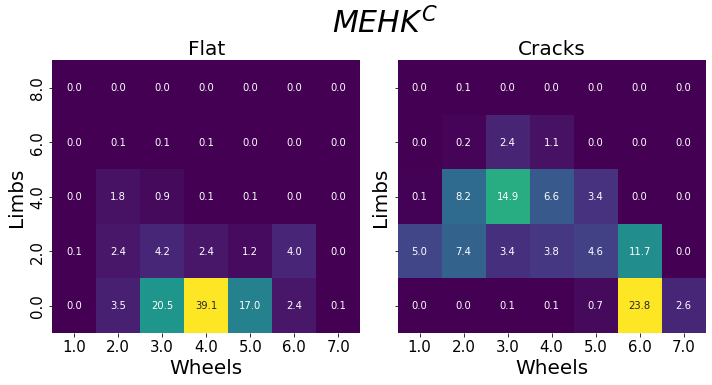} 
    \includegraphics[width=0.48\linewidth]{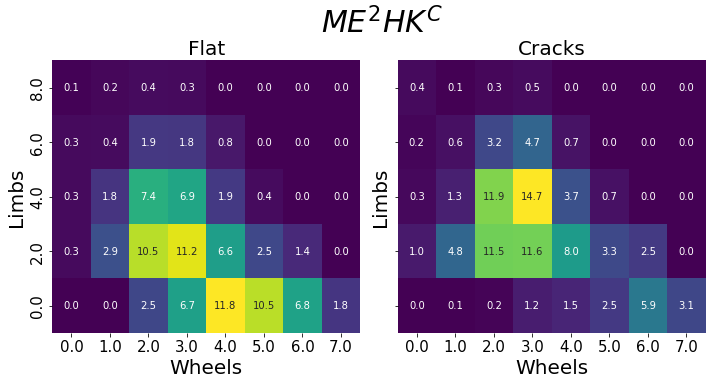} \\
    \includegraphics[width=0.48\linewidth]{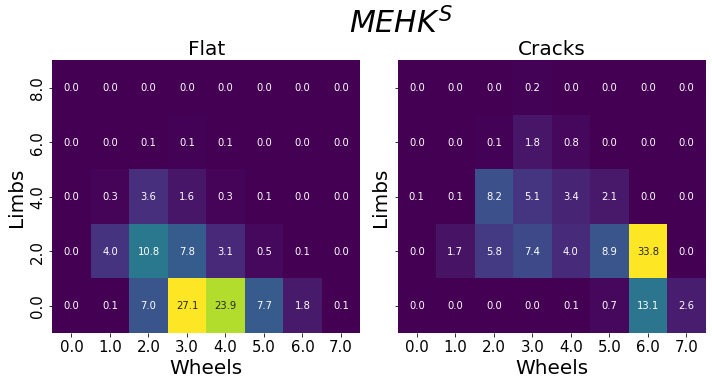} 
    \includegraphics[width=0.48\linewidth]{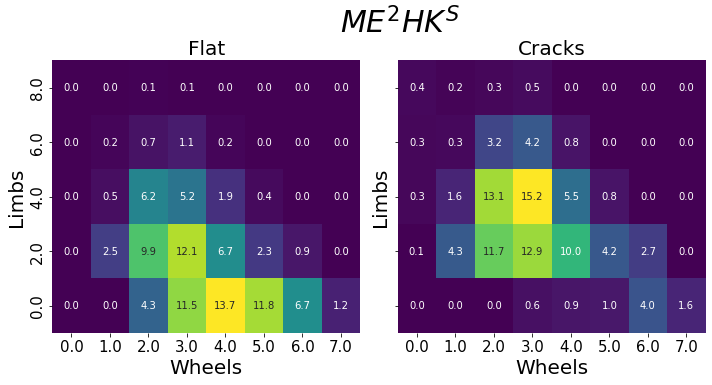}
    \caption{Distribution of the top 1\% robots over joints and wheels. From top to bottom: $MEHK$ - Dual CPPN, $ME^2HK$ - Dual CPPN, $MEHK$ - SQ-CPPN, $ME^2HK$ - SQ-CPPN}
    \label{fig:comp_distri}
    \vspace{-0.5cm}
\end{figure}

Figure~\ref{fig:comp_distri} focuses on the locomotion capability of the robots, showing the distribution of the number of robots w.r.t their number of wheels and limbs for the 100 robots with the highest exploration score aggregated over the 20 replicates, i.e. each plot displays a distribution of 2000 robots. For all the variants of $MEHK$, the distributions are different between the two environments. In the flat arena the majority of robots have wheels and one or no limbs. In the cracks environment, the majority of robots have between 1 and 3 limbs. 
This bias with regard to the terrain features is intuitive:  it is more efficient to locomote with wheels on a flat terrain and limbs are needed to avoid getting stuck in the cracks. MAP-Elites also helps increasing this difference as it explicitly search for high-performing and diverse robots.
The distribution of components is similar for both encodings, showing that it is an effect produced by the \textit{algorithm} itself. 

\begin{figure}[h!]
    \vspace{-0.3cm}
    \centering
    \includegraphics[width=\linewidth]{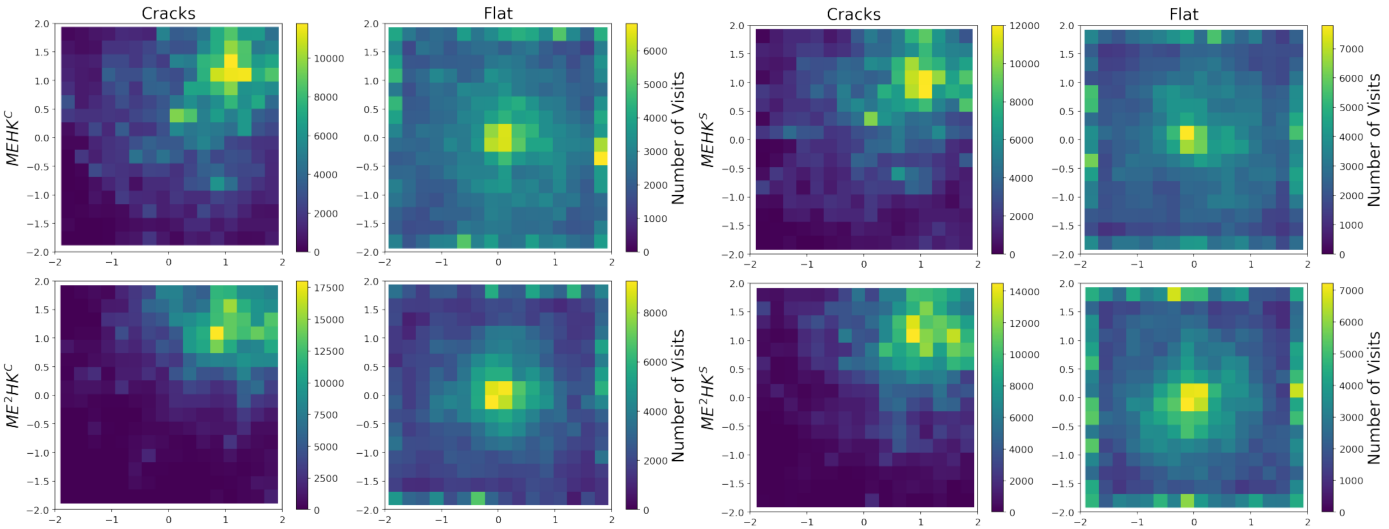} 
    \caption{Areas visited by the top 10 robots per replicates in term of exploration score using HKC. Each plot aggregates 200 robots trajectory.}
    \label{fig:trajs}
    \vspace{-0.5cm}

\end{figure}

Finally, Figure~\ref{fig:trajs} focuses on the behaviours of the robots. It shows the zones visited by the top 10 robots (in terms of exploration score) aggregated over 20 replicates (200 robots per plots). See Figure~\ref{fig:robex} for pictures of some these robots. The more yellow the square, the more robots visited this zone. As expected, the flat arena has a flared distribution centred on the starting position and with a peak on the border of the arena. This suggests that the robots are often moving around the arena by dragging along or staying stuck against the walls. For the cracks terrain, the distribution is also centred on the starting position (in the top right corner) with a circular spread. The centre of the distribution is more dense than the flat arena which is expected as a terrain  with cracks is harder than a flat terrain to navigate. These plots shows that the best robot generated have locomotion capabilities that can be potentially be optimised for downstream tasks. Also, it shows that if the robot design is well equipped for locomotion, e.g. having wheels and limbs in sensible positions, HKC is able to control the robots, confirming HKC is a good proxy to evaluate the mobility of robots.

\section{Conclusion}

We presented the novel algorithm $ME^2HK^S$ which generates diverse and functional robots using a superquadrics representation of the design of the robot chassis, while inducing diversity through MAP-Elites. The generated robots are diverse in term of the shape of their chassis and in the distribution of their active components, and functional in term of their ability to locomote.

The main findings can be summarised as follows. Firstly, we show that the use of the superquadrics representation to evolve the chassis improves results in terms of quality and diversity, regardless of the \textit{algorithm} used. This is because it has a smaller parameter space, and therefore, it is easier to navigate. And the superquadric equation is easier to interpret than a CPPN, allowing a tailored parametrization for our design space.  Secondly, using MAP-Elites as the algorithm that evolves robots improves quality and diversity regardless of the \textit{representation} used.  Finally, we demonstrate experimentally that an algorithm that incorporates both a superquadrics representation and MAP-Elites provides the best trade-off in terms of delivering a large archive of highly functional robots.
Finally, from a practical perspective, we remark that although MAP-Elites is well known to have low data-efficiency and often requires significant computational resources \cite{colas2020scaling,cully2021multi}, the use of the homeokinetic controller that quickly evaluates the locomotion capability of the robot enabled the discovery of nearly 4000 unique designs in a single run of $ME^2HK^S$.

Future work includes fine-tuning of robots for downstream tasks, and replacing the CPPN with a simpler and interpretable representation.

\footnotesize

\clearpage

\bibliographystyle{splncs04}
\bibliography{bib} 

@inproceedings{colas2020scaling,
  title={Scaling map-elites to deep neuroevolution},
  author={Colas, C{\'e}dric and Madhavan, Vashisht and Huizinga, Joost and Clune, Jeff},
  booktitle={Proceedings of the 2020 Genetic and Evolutionary Computation Conference},
  pages={67--75},
  year={2020}
}

@inproceedings{cully2021multi,
  title={Multi-emitter map-elites: improving quality, diversity and data efficiency with heterogeneous sets of emitters},
  author={Cully, Antoine},
  booktitle={Proceedings of the Genetic and Evolutionary Computation Conference},
  pages={84--92},
  year={2021}
}

@inproceedings{VeenstraEncodings, author = {Veenstra, Frank and Olsen, Martin Herring and Glette, Kyrre}, title = {Effects of encodings and quality-diversity on evolving 2D virtual creatures}, year = {2022}, isbn = {9781450392686}, publisher = {Association for Computing Machinery}, address = {New York, NY, USA}, url = {https://doi.org/10.1145/3520304.3529053}, doi = {10.1145/3520304.3529053}, booktitle = {Proceedings of the Genetic and Evolutionary Computation Conference Companion}, pages = {164–167}, numpages = {4}, location = {Boston, Massachusetts}, series = {GECCO '22} }

@article{han2023sq,
  title={Sq-slam: Monocular semantic slam based on superquadric object representation},
  author={Han, Xiao and Yang, Lu},
  journal={Journal of Intelligent \& Robotic Systems},
  volume={109},
  number={2},
  pages={29},
  year={2023},
  publisher={Springer}
}

@article{wu2025autonomous,
  title={Autonomous learning-free grasping and robot-to-robot handover of unknown objects},
  author={Wu, Yuwei and Li, Wanze and Liu, Zhiyang and Liu, Weixiao and Chirikjian, Gregory S},
  journal={Autonomous Robots},
  volume={49},
  number={3},
  pages={1--16},
  year={2025},
  publisher={Springer}
}

@inproceedings{vezzani2017grasping,
  title={A grasping approach based on superquadric models},
  author={Vezzani, Giulia and Pattacini, Ugo and Natale, Lorenzo},
  booktitle={2017 IEEE International Conference on Robotics and Automation (ICRA)},
  pages={1579--1586},
  year={2017},
  organization={IEEE}
}

@article{Dupac2012, author = {Dupac, Mihai}, title = {An object-oriented approach for mechanical components design and visualization}, year = {2012}, issue_date = {April 2012}, publisher = {Springer-Verlag}, address = {Berlin, Heidelberg}, volume = {28}, number = {2}, issn = {0177-0667}, journal = {Eng. with Comput.}, month = apr, pages = {95–107}, numpages = {13} }

@inproceedings{husbands1996two,
  title={Two applications of genetic algorithms to component design},
  author={Husbands, Phil and Jermy, Giles and McIlhagga, Malcolm and Ives, Robert},
  booktitle={AISB workshop on evolutionary computing},
  pages={50--61},
  year={1996},
  organization={Springer}
}

@inproceedings{xie2024map,
  title={A’MAP’to find high-performing soft robot designs: Traversing complex design spaces using MAP-elites and Topology Optimization},
  author={Xie, Yue and Pinskier, Josh and Liow, Lois and Howard, David and Iida, Fumiya},
  booktitle={2024 IEEE/RSJ International Conference on Intelligent Robots and Systems (IROS)},
  pages={11408--11415},
  year={2024},
  organization={IEEE}
}

@article{li2023evaluation,
  title={Evaluation of frameworks that combine evolution and learning to design robots in complex morphological spaces},
  author={Li, Wei and Buchanan, Edgar and Le Goff, L{\'e}ni K and Hart, Emma and Hale, Matthew F and Wei, Bingsheng and De Carlo, Matteo and Angus, Mike and Woolley, Robert and Gan, Zhongxue and others},
  journal={IEEE Transactions on Evolutionary Computation},
  volume={28},
  number={6},
  pages={1561--1574},
  year={2023},
  publisher={IEEE}
}

@inproceedings{auerbach2011, author = {Auerbach, Joshua E. and Bongard, Josh C.}, title = {Evolving complete robots with CPPN-NEAT: the utility of recurrent connections}, year = {2011}, isbn = {9781450305570}, publisher = {Association for Computing Machinery}, address = {New York, NY, USA}, url = {https://doi.org/10.1145/2001576.2001775}, doi = {10.1145/2001576.2001775}, booktitle = {Proceedings of the 13th Annual Conference on Genetic and Evolutionary Computation}, pages = {1475–1482}, numpages = {8}, location = {Dublin, Ireland}, series = {GECCO '11} }

@inproceedings{thomson2025understanding,
  title={Understanding the Navigation of Robot Morphology Spaces with Local Optima Network Analysis},
  author={Thomson, Sarah L and Le Goff, L{\'e}ni K and Hart, Emma and Eiben, Agoston E and Luck, Kevin S},
  booktitle={Artificial Life Conference Proceedings 37},
  volume={2025},
  number={1},
  pages={42},
  year={2025},
  organization={MIT Press One Rogers Street, Cambridge, MA 02142-1209, USA journals-info~…}
}

@inproceedings{mertan2025evolutionary,
  title={Evolutionary Brain-Body Co-Optimization Consistently Fails to Select for Morphological Potential},
  author={Mertan, Alican and Cheney, Nick},
  booktitle={Artificial Life Conference Proceedings 37},
  volume={2025},
  number={1},
  pages={63},
  year={2025},
  organization={MIT Press One Rogers Street, Cambridge, MA 02142-1209, USA journals-info~…}
}

@inproceedings{scarton23, author = {Scarton, Ludovico and Hagg, Alexander}, title = {On the Suitability of Representations for Quality Diversity Optimization of Shapes}, year = {2023}, isbn = {9798400701191}, publisher = {Association for Computing Machinery}, address = {New York, NY, USA}, url = {https://doi.org/10.1145/3583131.3590381}, doi = {10.1145/3583131.3590381}, booktitle = {Proceedings of the Genetic and Evolutionary Computation Conference}, pages = {963–971}, numpages = {9}, location = {Lisbon, Portugal}, series = {GECCO '23} }

@book{hesse2009self,
  title={Self-Organizing Control for Autonomous Robots A Dynamical Systems Approach Based on the Principle of Homeokinesis},
  author={Hesse, Frank},
  year={2009},
  publisher={Georg-August-Universitaet Goettingen (Germany)}
}

@article{cheney2018scalable,
  title={Scalable co-optimization of morphology and control in embodied machines},
  author={Cheney, Nick and Bongard, Josh and SunSpiral, Vytas and Lipson, Hod},
  journal={Journal of The Royal Society Interface},
  volume={15},
  number={143},
  pages={20170937},
  year={2018},
  publisher={The Royal Society}
}

@inproceedings{mertan2024investigating,
  title={Investigating premature convergence in co-optimization of morphology and control in evolved virtual soft robots},
  author={Mertan, Alican and Cheney, Nick},
  booktitle={European Conference on Genetic Programming (Part of EvoStar)},
  pages={38--55},
  year={2024},
  organization={Springer}
}

@article{xie2025morphology,
  title={The Morphology-Control Trade-Off: Insights into Soft Robotic Efficiency},
  author={Xie, Yue and Chu, Kai-fung and Wang, Xing and Iida, Fumiya},
  journal={arXiv preprint arXiv:2503.16127},
  year={2025}
}

@inproceedings{medvet2021biodiversity,
  title={Biodiversity in evolved voxel-based soft robots},
  author={Medvet, Eric and Bartoli, Alberto and Pigozzi, Federico and Rochelli, Marco},
  booktitle={Proceedings of the Genetic and Evolutionary Computation Conference},
  pages={129--137},
  year={2021}
}

@misc{mouret2015illuminating,
      title={Illuminating search spaces by mapping elites}, 
      author={Jean-Baptiste Mouret and Jeff Clune},
      year={2015},
      eprint={1504.04909},
      archivePrefix={arXiv},
      primaryClass={cs.AI},
      url={https://arxiv.org/abs/1504.04909}, 
}

@article{buchanan2020bootstrapping,
  title={Bootstrapping artificial evolution to design robots for autonomous fabrication},
  author={Buchanan, Edgar and Le Goff, L{\'e}ni K and Li, Wei and Hart, Emma and Eiben, Agoston E and De Carlo, Matteo and Winfield, Alan F and Hale, Matthew F and Woolley, Robert and Angus, Mike and others},
  journal={Robotics},
  volume={9},
  number={4},
  pages={106},
  year={2020},
  publisher={MDPI}
}

@article{gupta2021embodied,
  title={Embodied intelligence via learning and evolution},
  author={Gupta, Agrim and Savarese, Silvio and Ganguli, Surya and Fei-Fei, Li},
  journal={Nature communications},
  volume={12},
  number={1},
  pages={5721},
  year={2021},
  publisher={Nature Publishing Group UK London}
}

@article{cheney2014unshackling,
  title={Unshackling evolution: evolving soft robots with multiple materials and a powerful generative encoding},
  author={Cheney, Nick and MacCurdy, Robert and Clune, Jeff and Lipson, Hod},
  journal={ACM SIGEVOlution},
  volume={7},
  number={1},
  pages={11--23},
  year={2014},
  publisher={ACM New York, NY, USA}
}

@article{kriegman2020scalable,
  title={A scalable pipeline for designing reconfigurable organisms},
  author={Kriegman, Sam and Blackiston, Douglas and Levin, Michael and Bongard, Josh},
  journal={Proceedings of the National Academy of Sciences},
  volume={117},
  number={4},
  pages={1853--1859},
  year={2020},
  publisher={National Acad Sciences}
}

@incollection{lindenmayer1992grammars,
  title={Grammars of development: discrete-state models for growth, differentiation, and gene expression in modular organisms},
  author={Lindenmayer, Aristid and J{\"u}rgensen, Hans},
  booktitle={Lindenmayer systems: Impacts on theoretical computer science, computer graphics, and developmental biology},
  pages={3--21},
  year={1992},
  publisher={Springer}
}

@article{angus2023practical,
  title={Practical hardware for evolvable robots},
  author={Angus, Mike and Buchanan, Edgar and Le Goff, L{\'e}ni K and Hart, Emma and Eiben, Agoston E and De Carlo, Matteo and Winfield, Alan F and Hale, Matthew F and Woolley, Robert and Timmis, Jon and others},
  journal={Frontiers in Robotics and AI},
  volume={10},
  year={2023},
  publisher={Frontiers Media SA}
}

@article{homeok,
title = "Homeokinetic approach to autonomous learning in mobile robots",
author = "R Der and JM Herrmann and R Liebscher",
year = "2002",
language = "English",
volume = "1679",
pages = "301--308",
journal = "VDI BERICHTE",
publisher = "VDI; 1999",
}

@article{stanley2007compositional,
  title={Compositional pattern producing networks: A novel abstraction of development},
  author={Stanley, Kenneth O},
  journal={Genetic programming and evolvable machines},
  volume={8},
  pages={131--162},
  year={2007},
  publisher={Springer}
}

@inproceedings{miras2018effects,
  title={Effects of selection preferences on evolved robot morphologies and behaviors},
  author={Miras, Karine and Haasdijk, Evert and Glette, Kyrre and Eiben, AE},
  booktitle={Artificial Life Conference Proceedings},
  pages={224--231},
  year={2018},
  organization={MIT Press One Rogers Street, Cambridge, MA 02142-1209, USA}
}

@article{miras2021constrained,
  title={Constrained by design: Influence of genetic encodings on evolved traits of robots},
  author={Miras, Karine},
  journal={Frontiers in Robotics and AI},
  volume={8},
  pages={672379},
  year={2021},
  publisher={Frontiers Media SA}
}

@inproceedings{le2020pros,
  title={On pros and cons of evolving topologies with novelty search},
  author={Le Goff, L{\'e}ni K and Hart, Emma and Coninx, Alexandre and Doncieux, St{\'e}phane},
  booktitle={Artificial Life Conference Proceedings},
  pages={423--431},
  year={2020},
  organization={MIT Press One Rogers Street, Cambridge, MA 02142-1209, USA}
}

@INPROCEEDINGS{LeGoffICRA24,
  author={Goff, Leni K. Le and Smith, Simón C.},
  booktitle={2025 IEEE International Conference on Robotics and Automation (ICRA)}, 
  title={Efficient and Diverse Generative Robot Designs Using Evolution and Intrinsic Motivation}, 
  year={2025},
  volume={},
  number={},
  pages={16598-16605},
  doi={10.1109/ICRA55743.2025.11127289}}

@article{legoff2024efficient,
  title={Efficient and Diverse Generative Robot Designs using Evolution and Intrinsic Motivation},
  author={Le Goff, Leni K and Smith, Simon C},
  journal={arXiv preprint arXiv:2411.18423},
  year={2024}
}

@article{nordmoen2021map,
  title={Map-elites enables powerful stepping stones and diversity for modular robotics},
  author={Nordmoen, J{\o}rgen and Veenstra, Frank and Ellefsen, Kai Olav and Glette, Kyrre},
  journal={Frontiers in Robotics and AI},
  volume={8},
  pages={639173},
  year={2021},
  publisher={Frontiers Media SA}
}

@inproceedings{Nadizar25, author = {Nadizar, Giorgia and Rusin, Francesco and Medvet, Eric and Ochoa, Gabriela}, title = {The Role of Stepping Stones in MAP-Elites: Insights from Search Trajectory Networks}, year = {2025}, isbn = {978-3-031-89990-4}, publisher = {Springer-Verlag}, address = {Berlin, Heidelberg}, url = {https://doi.org/10.1007/978-3-031-89991-1_14}, doi = {10.1007/978-3-031-89991-1_14}, booktitle = {Genetic Programming: 28th European Conference, EuroGP 2025, Held as Part of EvoStar 2025, Trieste, Italy, April 23–25, 2025, Proceedings}, pages = {224–239}, numpages = {16}, location = {Trieste, Italy} }

@book{der2012playful,
  title={The playful machine: theoretical foundation and practical realization of self-organizing robots},
  author={Der, Ralf and Martius, Georg},
  volume={15},
  year={2012},
  publisher={Springer Science \& Business Media}
}

@article{soodak1978homeokinetics,
  title={Homeokinetics: A physical science for complex systems},
  author={Soodak, Harry and Iberall, Arthur},
  journal={Science},
  volume={201},
  number={4356},
  pages={579--582},
  year={1978},
  publisher={American Association for the Advancement of Science}
}

@inproceedings{le2024improving,
  title={Improving Efficiency of Evolving Robot Designs via Self-Adaptive Learning Cycles and an Asynchronous Architecture},
  author={Le Goff, Leni and Hart, Emma},
  booktitle={Proceedings of the Genetic and Evolutionary Computation Conference Companion},
  pages={1607--1615},
  year={2024}
}

@inproceedings{smith2011homeokinetic,
  title={Homeokinetic reinforcement learning},
  author={Smith, Sim{\'o}n C and Herrmann, J Michael},
  booktitle={IAPR International Workshop on Partially Supervised Learning},
  pages={82--91},
  year={2011},
  organization={Springer}
}

\end{document}